\pdfoutput=1

\documentclass[11pt]{article}

\usepackage[final]{acl}
\usepackage[utf8]{inputenc} 

\usepackage[T1]{fontenc}    
\usepackage{hyperref}       
\usepackage{url}            
\usepackage{booktabs}       
\usepackage{amsfonts}       
\usepackage{amsmath}        
\usepackage{nicefrac}       
\usepackage{microtype}      
\usepackage[table]{xcolor}         
\usepackage{multirow}
\usepackage{tcolorbox}

\usepackage{graphicx}
\usepackage{longtable}
\usepackage{caption}
\usepackage{geometry}
\usepackage{adjustbox}
\usepackage{float}      

\usepackage{colortbl}  
\usepackage{array}     
\usepackage{amsmath}   
\usepackage{booktabs}  
\usepackage{makecell}  
\usepackage{wrapfig}   

\usepackage{graphicx}
\usepackage{subfigure}
\usepackage{url}
\usepackage{arydshln}
\usepackage{bm}

\usepackage{times}
\usepackage{latexsym}

\usepackage[T1]{fontenc}

\usepackage[utf8]{inputenc}

\usepackage{microtype}

\usepackage{inconsolata}

\usepackage{graphicx}

\title{Instructions are all you need: Self-supervised Reinforcement Learning for Instruction Following}

\author{
 \textbf{Qingyu Ren\textsuperscript{1}\thanks{\ Equal contribution.}},
 \textbf{Qianyu He\textsuperscript{1}\footnotemark[1]},
 \textbf{Powei Chang\textsuperscript{2}}, \textbf{Jie Zeng\textsuperscript{1}},
 \textbf{Zeye Sun\textsuperscript{3}}, \textbf{Fei Yu\textsuperscript{3}},\\
 \textbf{Jiaqing Liang\textsuperscript{2}\thanks{\ Corresponding author.}},
 \textbf{Yanghua Xiao\textsuperscript{1}\footnotemark[2]}\\
    \textsuperscript{\rm 1}Shanghai Key Laboratory of Data Science, \\College of Computer Science and Artificial Intelligence, Fudan University,\\
    \textsuperscript{\rm 2}School of Data Science, Fudan University,
    \textsuperscript{\rm 3}Ant Group\\
     \{qyren24,qyhe21,bwzhang24, jzeng23\}@m.fudan.edu.cn, \{liangjiaqing, shawyh\}@fudan.edu.cn
}

\begin{document}
\maketitle
\begin{abstract}

Language models often struggle to follow multi-constraint instructions that are crucial for real-world applications.
Existing reinforcement learning (RL) approaches suffer from dependency on external supervision and sparse reward signals from multi-constraint tasks.
We propose a label-free self-supervised RL framework that eliminates dependency on external supervision by deriving reward signals directly from instructions  and generating pseudo-labels for reward model training.
Our approach introduces constraint decomposition strategies and efficient constraint-wise binary classification to address sparse reward challenges while maintaining computational efficiency.
Experiments show that our approach generalizes well, achieving strong improvements across 3 in-domain and 5 out-of-domain datasets, including challenging agentic and multi-turn instruction following. The code and data are publicly available at \href{https://github.com/Rainier-rq/verl-if}{https://github.com/Rainier-rq/verl-if} and \href{https://huggingface.co/dd12345789}{https://huggingface.co/dd12345789}.
\end{abstract}

\section{Introduction}

Instruction following capabilities (i.e., the ability to follow multiple constraints simultaneously) are crucial to ensure practical use of language models in real-world scenarios~\cite{zhang2025recommendation,li2025thinking}.
On one hand, real-world conversations with human users often contain multiple constraints in the instructions~\cite{deshpande2025multichallenge,wen2024benchmarking}.
On the other hand, reliable instruction following is essential for models in complex agentic tasks~\cite{qi2025agentif}.
However, instruction-tuned models often demonstrate inadequate adherence to user instructions.
Meanwhile, reasoning models, though specifically trained for  various reasoning abilities~\cite{o1, guo2025deepseek, seed2025seed1}, tend to sacrifice instruction-following abilities.
\begin{figure}[t] 
    \centering
            \includegraphics[width=0.5\textwidth]{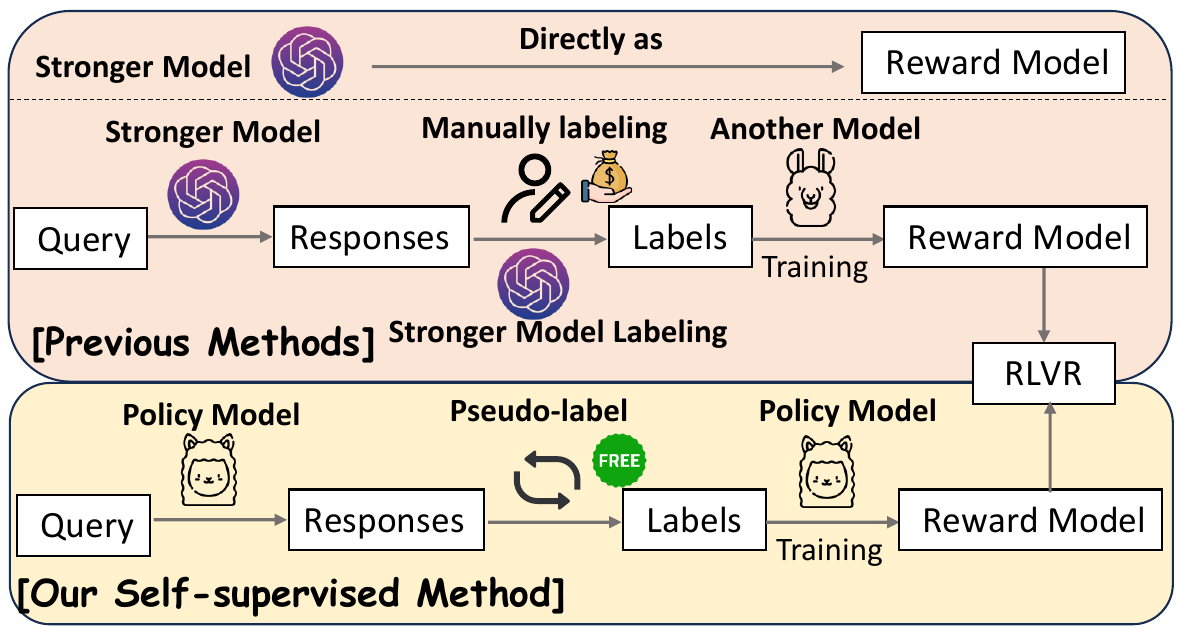}
    \caption{Comparison between our self-supervised RL method and previous methods. Our method does not rely on external sources to generate outputs or labels, which is both effective and efficient.}
    \label{fig:motivation}
\end{figure}


\begin{figure*}[t] 
    \centering
            \includegraphics[width=0.9\textwidth]{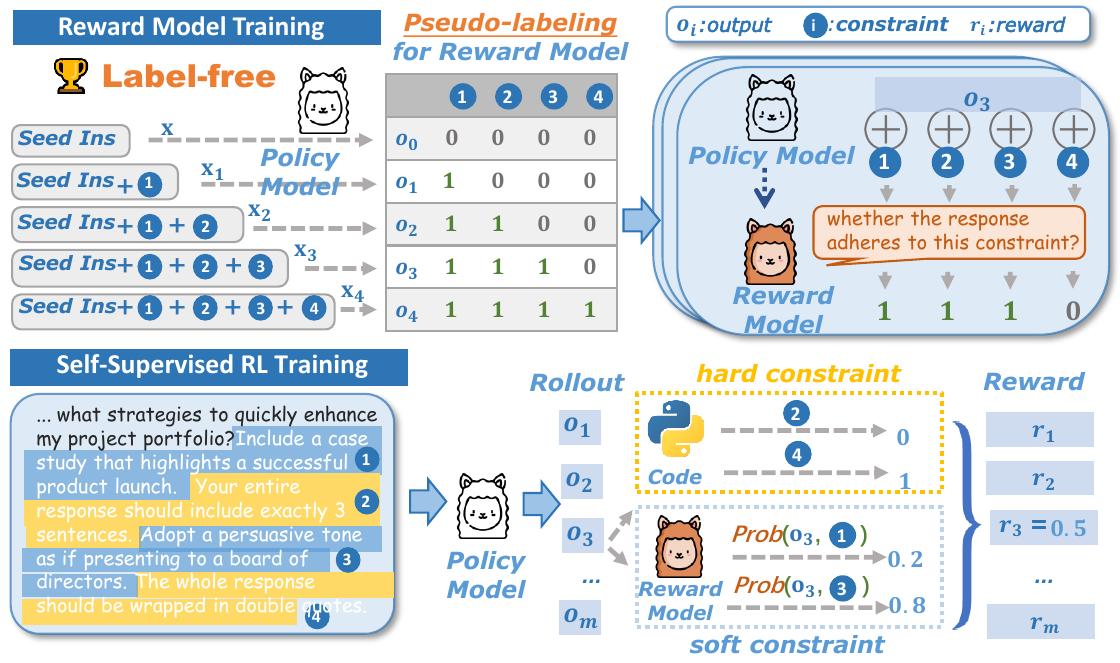}
    \caption{Overview of our self-supervised RL framework: from label-free instructions, we generate pseudo-labels for constraint-wise reward model training and then train the policy model with both hard constraints via rule-based verification and soft constraints via the reward model.}
    \label{fig:framework}
\end{figure*}
Existing methods for improving instruction-following abilities focus on obtaining high-quality external supervision signals and outputs for training~\cite{qin2025incentivizing,ren2025stepbystepmasteryenhancingsoft}. These methods typically employ supervised fine-tuning (SFT) with distilled data from stronger models~\cite{sun2024conifer} or direct preference optimization (DPO) with collected pairwise preference data~\cite{he2024complex,huang2025musc}.
However, reinforcement learning with verifiable rewards (RLVR) presents a more suitable paradigm for multi-constraint instruction following tasks.
First, RLVR only requires instructions with verifiable reward signals, rather than high-quality outputs from external models. Moreover, 
constraints in complex instructions are inherently verifiable~\cite{yang2025qwen3,peng2025verif}, making RLVR  suited for our framework.

Existing RLVR approaches have following limitations. 
First, as shown in Fig.~\ref{fig:motivation}, the supervision signal problem: existing approaches rely heavily on stronger models, either directly as reward models or as sources for distilling reward model training data, or on manual data annotation~\cite{qin2025incentivizing}.
Stronger models are often proprietary or computationally expensive, and their inherent ability  can also restrict the improvement of other models.
Moreover, the sparse reward signal problem: multi-constraint instructions are  difficult to satisfy completely, leading to sparse  rewards that hinder effective learning~\cite{yu2025dapo}.
Furthermore, generative reward models are computationally heavy, causing slow training, particularly for tasks with multiple constraints~\cite{yu2025rewardanything}.

To address these challenges, we propose an efficient self-supervised RL framework that enhances instruction following capabilities using only label-free instructions as inputs. 
Specifically, to tackle the supervision signal challenge, we develop a pseudo-label generation mechanism that the model can derive reward signals directly from instructions, achieving true label-free training without external dependencies.
By label-free, we emphasize that our approach requires no human annotations, preference datasets, or external reward models—the model generates all supervision signals autonomously from the instructions themselves.
For the sparse reward issue, we introduce  a multi-constraint decomposition strategy that breaks down multi-constraint instructions into simpler sub-tasks with incrementally increasing complexity, establishing dense learning signals throughout the training process~\cite{yu2025dapo}.
Additionally, we design an efficient constraint-wise binary classification approach that evaluates each constraint independently, ensuring computational efficiency while maintaining high effectiveness.
Our framework shows remarkable generalization abilities, showing consistent improvements across 3 in-domain and 5 out-of-domain datasets, including challenging  agentic and multi-turn instruction following.

Overall, our contributions are as follows:
(1) We propose a self-supervised RL framework that enhances instruction following capabilities using only label-free instructions as inputs.
(2) We design an efficient and accurate reward modeling approach that handles multi-constraint scenarios through constraint-wise binary classification.
(3) Extensive experiments demonstrate that our trained models achieve significant improvements in instruction following abilities with strong generalization to
out-of-domain tasks, while maintaining general capabilities.

\section{Related Work}
\subsection{Complex Instruction Following Improvement}
Some works distill responses from stronger models through supervised fine-tuning~\cite{sun2024conifer, qin2025incentivizing} or collecting pairwise preference data for direct preference optimization~\cite{he2024complex, qi2024constraint}.
Others adopt self-play approaches that enhance model capabilities through code generation verification~\cite{dong2024self} or training additional refiner models~\cite{cheng2024spar}.
Unlike these methods, we do not rely on stronger models and instead train exclusively through reinforcement learning based on the model's own abilities.

\subsection{Reinforcement Learning for Complex Instruction Following}
Some works use rule-based rewards for hard constraints~\cite{pyatkin2025generalizing}, but these methods cannot generalize to soft constraints.
Other works either rely on a stronger model as the reward model or utilize the stronger model to distill data for reward model training~\cite{peng2025verif}.
Additionally, existing approaches often suffer from computational inefficiency.
Different from these works, our reward model is independent of stronger reasoning models, requires no distillation, and operates more efficiently.

\section{Method}

Our framework consists of two main stages. First, as illustrated in Fig.~\ref{fig:framework}, we construct a multi-constraint instruction dataset and decompose complex instructions into incremental constraint curricula to provide dense training signals. 
We then train constraint-wise binary classification reward models using pseudo-labels generated directly from the instructions. Second, we then apply GRPO~\cite{shao2024deepseekmath} to optimize the policy model using the composite reward signals.

\subsection{Dataset Construction}

Our training dataset consists of two main components: (1) synthetically generated multi-constraint instructions, and (2) general reasoning data from math and science domains to maintain the model's overall capabilities. 

\paragraph{Complex Instruction Synthesis.}\label{sec:constuct}

To ensure diversity in our multi-constraint instruction dataset, we cover both hard and soft constraint types~\cite{ren2025stepbystepmasteryenhancingsoft}. 
We begin by collecting a diverse set of 3,000 seed instructions from different sources~\cite{kopf2024openassistant,wang2022self,wang2022super,li2025infinity}. Subsequently, we use GPT-4o to add multiple constraints to these seed instructions. Our self-supervision supervises the outputs and labels, not the instructions. For hard constraints, we include 23 types, such as JSON format and frequency of all-capital words. For soft constraints, we include 25 types, such as style and role-based constraints. Details of the constraint types can be found in Appx.~\ref{appx:ct}. 


\paragraph{General Reasoning Data Integration.}

To maintain the general abilities of the model, we integrate math and science data into our dataset. Specifically, we select 4,501 math problems from the DeepScaleR-Preview-Dataset~\cite{luo2025deepscaler} and 1,929 science questions from SciKnowEval~\cite{feng2024sciknoweval}. 

\paragraph{Incremental Constraint Curriculum.}\label{sec:const}

Complex instruction-following is challenging~\cite{zhang2024cfbench}, leading to sparse reward signals during RL training~\cite{yu2025dapo}. We decompose complex instructions for progressive learning.
Given a multi-constraint instruction $x$ with constraints $\{c_1 \oplus  c_2 \oplus  \ldots \oplus c_n\}$, we create curriculum levels $\mathcal{L}_k$ where each level $k$ contains sub-instruction $x_k$ with the first $k$ constraints:
$
x_k = x \text{ with constraints } \{c_1, c_2, \ldots, c_k\}
$.
This creates a progressive curriculum from single-constraint ($\mathcal{L}_1$) to full multi-constraint instructions ($\mathcal{L}_n$). Statistical details are in Tab.~\ref{tab:num}. 

\newcolumntype{g}{>{\columncolor{green!10}}c}
\setlength\tabcolsep{7pt}
\begin{table}[t]
\centering
\huge
\newcolumntype{b}{>{\columncolor{blue!10}}c}
\renewcommand{\arraystretch}{1.1}
\resizebox{0.4\textwidth}{!}{
\begin{tabular}{cccccc}
\toprule
 Curriculum    & \# Instruct.  &\# Cons.         & \# Soft. & \#Hard.   \\ \midrule
$\mathcal{L}_1$                   &   2806       &2806&1578&1228\\
$\mathcal{L}_2$ &2745&5490 &3203  &2287   \\ 
$\mathcal{L}_3$  &2700     &8100& 4833&3267   \\
$\mathcal{L}_4$          &  2700     &10800   & 6481&4319 \\
$\mathcal{L}_5$              &  2619    &13095 & 8101 &4994   \\ 
\bottomrule
\end{tabular}
}
 \caption{Statistics of curricula. \#Instruct, \#Constraints, \#Soft, and \#Hard refer to the number of instruction constraints, total constraints, soft constraints, and hard constraints, respectively. 
}
 \label{tab:num}
\end{table}
\subsection{Reward Modeling}

To model constraint satisfaction, we design different reward mechanisms for hard and soft constraints to produce constraint-level rewards. 

\paragraph{Hard Constraint Modeling.}
For hard constraints that can be directly verified using explicit rules~\cite{pyatkin2025generalizing}, we adopt programmatic verification. For an input example $(o, c)$ with response $o$ and constraint $c$, we define a binary constraint-level reward function:
\[
R_{h}(o, c) = 
\begin{cases}
1, & \text{if } o \text{ satisfies constraint }  c \\
0, & \text{otherwise}
\end{cases}
\]


\begin{table}[t]
\centering
\renewcommand{\arraystretch}{1.0}
\resizebox{0.9\columnwidth}{!}{%
\begin{tabular}{lccc}
\toprule
\multicolumn{1}{l}{} & \multicolumn{1}{l}{Kendall Tau Coefficient}&\multicolumn{1}{l}{ Position Consistency}\\ \midrule
Our  Dataset                      &  94.0    &  97.0              \\ 
\bottomrule
\end{tabular}
}
\caption{
Agreement between constructed dataset and human annotation. The detailed evaluation setup is provided in  Appx.~\ref{appx:es}}
  \label{tab:human_label}
\end{table}
\paragraph{Soft Constraint Modeling.}
To model soft constraints that cannot be verified through rules, avoid external supervision, and achieve efficiency, we train a binary classification reward model using pseudo-labels constructed from the instructions themselves without external labels.

During constraint decomposition, a natural relationship emerges: for constraint $c_k$, the response $o_k$ (generated for instruction with constraint $c_k$) exhibits stronger adherence to the constraint than $o_{k-1}$ (generated for instruction without $c_k$). This allows us to construct pseudo-labels: (1) \textit{Positive sample} $(o_k, c_k, label=1)$: response satisfies the constraint, 
(2) \textit{Negative sample} $(o_{k-1}, c_k, label=0)$: response does not satisfy the constraint.
We define a constraint-level reward function $f(o, c) \rightarrow [0, 1]$ that estimates the probability that response $o$ satisfies constraint $c$. The model is trained to minimize the Binary Cross-Entropy loss:

\begin{small}
\[
\mathcal{L} = - \sum_{k=1}^n \left[ \log f(o_k, c_k) + \log \big(1 - f(o_{k-1}, c_k)\big) \right].
\]
\end{small}
As shown in Tab.~\ref{tab:human_label}, our self-supervised dataset demonstrates high consistency with humans. 


\subsection{RL Training}

With the constraint-level reward signals established, we introduce how to utilize these models during reinforcement learning training to predict sample-level rewards for policy optimization.

\paragraph{Reward Model Usage During Training.}
For soft constraints, the trained reward model $f(o, c)$ takes a response $o$ and constraint $c$ as input, producing logits over two classes (satisfy/not satisfy). These logits are converted to probabilities:

\begin{small}
\[
R_s(o, c) = \frac{\exp(logits[1])}
         {\exp(logits[0]) + \exp(logits[1])}
\]
\end{small}

This gives a scalar reward value $R_s(o, c) \in [0, 1]$ representing the probability that response $o$ satisfies soft constraint $c$. 

\paragraph{Constraint-Level Reward Prediction.}
We aggregate constraint-level rewards into sample-level rewards. For instruction $x_k$ with constraints $\{c_1 \oplus c_2, \ldots, c_k\}$ and policy-generated response $o_k$, the sample-level reward is:

\begin{small}
$$
R_f = \frac{1}{k} \sum_{i=1}^{k} r_i, \quad r_i =
\begin{cases}
R_s(o_k, c_i), & \text{if } c_i \text{ is soft} \\
R_h(o_k, c_i), & \text{if } c_i \text{ is hard}
\end{cases}
$$
\end{small}
For reasoning tasks, we assign $R_f = 1$ for correct answers and $R_f = 0$ otherwise.
This composite reward $R_f \in [0, 1]$ captures the overall constraint satisfaction of the response and serves as the reward signal for GRPO optimization. 


\section{Experiment}

\definecolor{steelblue}{RGB}{70,130,180}
\definecolor{lightgray}{gray}{0.93}

\begin{table*}[t]
\centering
\renewcommand{\arraystretch}{1.3}
\resizebox{\textwidth}{!}{
\begin{tabular}{l!{\vrule width 1pt}c!{\vrule width 1pt}ccc|ccccc}
\toprule
\multirow{3}{*}{\textbf{Models}} & \multirow{3}{*}{\textbf{Base Model}} 
& \multicolumn{3}{c|}{\cellcolor{lightgray}\textbf{In-Domain}} 
& \multicolumn{5}{c}{\cellcolor{lightgray}\textbf{Out-of-Domain}} \\ 
\cmidrule(lr){3-5} \cmidrule(lr){6-10}
& & \textbf{IFEval} & \textbf{CFBench} & \textbf{FollowBench} & \textbf{ComplexBench} & \textbf{WritingBench} & \textbf{Collie} & \textbf{AgentIF} & \textbf{MultiChallenge}  \\ 
\cmidrule(lr){3-5} \cmidrule(lr){6-10}
& & \textbf{Pr.(L)} & \textbf{ISR} & \textbf{HSR} & \textbf{Overall} & \textbf{Avg.} & \textbf{Avg.} & \textbf{CSR} & \textbf{Overall} \\ 
\midrule
GPT-4o&GPT&84.8&65.3&70.4&71.6&7.5&49.8&58.5&12.9\\
QwQ-32B&Qwen-32B&83.9&68.0&62.2&73.3&7.9&52.4&58.1&38.5\\
IR-1.5B & Qwen-1.5B & 57.7 & 22.0 & 37.8 & 44.4 & 4.0 & 16.3 & 38.7 & 12.5 \\ 
Conifer-7B-DPO & Mistral-7B & 52.3 & 25.0 & 50.0 & 48.1 & 3.2 & 17.8 & 44.3 & 8.0\\  
Crab-7B-DPO & Mistral-7B & 57.7 & 25.0 & 49.4 & 59.0 & 4.5 & 19.6 & 47.2 & 14.1 \\  
SPAR-8B-DPO & LLaMA-8B & 82.4& 37.0 & 56.1 & 63.8 & 4.7 & 27.7 & 53.6 & 17.1\\  
VERIF-8B & LLaMA-8B & 87.1 & 41.0 & 56.9 & 54.7 & 5.1 & 28.3 & 56.6 & 15.0\\  
\midrule
Qwen2.5-1.5B-Instruct & Qwen-1.5B & 43.6 & 22.0 & 34.6 & 45.9 & 4.5 & 13.0 & 42.8 & 12.0\\  
\rowcolor{steelblue!25}
Qwen2.5-1.5B-Instruct-IF & Qwen-1.5B & 
65.2\textcolor{red}{\scriptsize (+21.6)} & 
29.0\textcolor{red}{\scriptsize (+7.0)} & 
39.0\textcolor{red}{\scriptsize (+4.4)} & 
48.6\textcolor{red}{\scriptsize (+2.7)} & 
4.6\textcolor{red}{\scriptsize (+0.1)} & 
16.7\textcolor{red}{\scriptsize (+3.7)} & 
48.2\textcolor{red}{\scriptsize (+5.4)} & 
13.3\textcolor{red}{\scriptsize (+1.3)}\\  
Qwen2.5-7B-Instruct & Qwen-7B & 73.9 & 47.0 & 55.1 & 66.1 & 5.7 & 36.3 & 54.2 & 15.2 \\  
\rowcolor{steelblue!25}
Qwen2.5-7B-Instruct-IF & Qwen-7B & 
78.9\textcolor{red}{\scriptsize (+5.0)} & 
52.0\textcolor{red}{\scriptsize (+5.0)} & 
57.5\textcolor{red}{\scriptsize (+2.4)} & 
68.7\textcolor{red}{\scriptsize (+2.6)} & 
5.8\textcolor{red}{\scriptsize (+0.1)} & 
38.0\textcolor{red}{\scriptsize (+1.7)} & 
56.7\textcolor{red}{\scriptsize (+2.5)} & 
15.6\textcolor{red}{\scriptsize (+0.4)}\\  
Distill-Qwen-1.5B & Qwen-1.5B & 42.3 & 17.0 & 22.7 & 39.8 & 3.9 & 14.0 & 37.5 & 12.0\\  
\rowcolor{steelblue!25}
Distill-Qwen-1.5B-IF & Qwen-1.5B & 
58.8\textcolor{red}{\scriptsize (+16.5)} & 
20.0\textcolor{red}{\scriptsize (+3.0)} & 
26.4\textcolor{red}{\scriptsize (+3.7)} & 
43.3\textcolor{red}{\scriptsize (+3.5)} & 
4.1\textcolor{red}{\scriptsize (+0.2)} & 
14.5\textcolor{red}{\scriptsize (+0.5)} & 
39.7\textcolor{red}{\scriptsize (+2.2)} & 
13.1\textcolor{red}{\scriptsize (+1.1)}\\  
Distill-Qwen-7B & Qwen-7B & 61.7 & 36.0 & 41.7 & 55.2 & 5.3 & 25.2 & 47.2 & 13.9\\  
\rowcolor{steelblue!25}
Distill-Qwen-7B-IF & Qwen-7B & 
71.7\textcolor{red}{\scriptsize (+10.0)} & 
42.0\textcolor{red}{\scriptsize (+6.0)} & 
49.1\textcolor{red}{\scriptsize (+7.4)} & 
60.7\textcolor{red}{\scriptsize (+5.5)} & 
5.6\textcolor{red}{\scriptsize (+0.3)} & 
27.0\textcolor{red}{\scriptsize (+2.0)} & 
46.9\textcolor{gray}{\scriptsize (-0.3)} & 
16.3\textcolor{red}{\scriptsize (+2.4)}\\  
Llama-3.1-8B-Instruct & LLaMA-8B & 73.8 & 34.0 & 53.8 & 63.6 & 4.7 & 46.5 & 53.4 & 16.2  \\  
\rowcolor{steelblue!25}
Llama-3.1-8B-Instruct-IF & LLaMA-8B & 
83.0\textcolor{red}{\scriptsize (+9.2)} & 
40.0\textcolor{red}{\scriptsize (+6.0)} & 
57.5\textcolor{red}{\scriptsize (+3.7)} & 
64.5\textcolor{red}{\scriptsize (+0.9)} & 
4.8\textcolor{red}{\scriptsize (+0.1)} & 
50.2\textcolor{red}{\scriptsize (+3.7)} & 
55.1\textcolor{red}{\scriptsize (+1.7)} & 
19.1\textcolor{red}{\scriptsize (+2.9)} \\  
Distill-Qwen-14B&Qwen-14B&74.9&55.0&51.2&72.7&6.0&34.4&54.5&17.2\\
\rowcolor{steelblue!25}
Distill-Qwen-14B-IF&Qwen-14B&
83.7\textcolor{red}{\scriptsize (+8.8)}&
64.0\textcolor{red}{\scriptsize (+9.0)}&
57.6\textcolor{red}{\scriptsize (+6.4)}&
76.9\textcolor{red}{\scriptsize (+4.2)}&
6.4\textcolor{red}{\scriptsize (+0.4)}&
37.6\textcolor{red}{\scriptsize (+3.2)}&
57.0\textcolor{red}{\scriptsize (+2.5)}&
26.8\textcolor{red}{\scriptsize (+9.6)}\\
R1-0528-Qwen3-8B & Qwen-8B & 79.7 & 66.0 & 60.4 & 68.5 & 7.6 & 36.9 & 57.4 & 21.4 \\  
\rowcolor{steelblue!25}
R1-0528-Qwen3-8B & Qwen-8B & 
87.1\textcolor{red}{\scriptsize (+7.4)} & 
68.0\textcolor{red}{\scriptsize (+2.0)} & 
63.8\textcolor{red}{\scriptsize (+3.4)} & 
71.1\textcolor{red}{\scriptsize (+2.6)} & 
7.1\textcolor{gray}{\scriptsize (-0.5)} & 
37.1\textcolor{red}{\scriptsize (+0.2)} & 
63.5\textcolor{red}{\scriptsize (+6.1)} & 
28.7\textcolor{red}{\scriptsize (+7.3)}\\  
Qwen2.5-32B-Instruct&Qwen-32B& 82.4&59.0&65.1&77.7&5.5&43.3&64.9&17.3\\
\rowcolor{steelblue!25}
Qwen2.5-32B-Instruct-IF&Qwen-32B&
85.0\textcolor{red}{\scriptsize (+2.6)}&
63.0\textcolor{red}{\scriptsize (+4.0)}&
66.8\textcolor{red}{\scriptsize (+1.7)}&
79.8\textcolor{red}{\scriptsize (+2.1)}&
5.6\textcolor{red}{\scriptsize (+0.1)}&
48.1\textcolor{red}{\scriptsize (+4.8)}&
66.9\textcolor{red}{\scriptsize (+2.0)}&
21.3\textcolor{red}{\scriptsize (+4.0)}\\
\bottomrule
\end{tabular}
}
\caption{
Overall performance on In-Domain and Out-of-Domain benchmarks. Pr.(L) denotes loose prompt.
}
\label{tab:main}
\end{table*}

\subsection{Experiment Setup}

We experiment on both non-reasoning models (Qwen2.5-1.5B/7B/32B-Instruct, Llama-3.1-8B-Instruct) and reasoning models distilled from R1 (DeepSeek-R1-Distill-Qwen-1.5B/7B/14B and DeepSeek-R1-0528-Qwen3-8B). \textbf{IF} denotes models trained with our method. We select several models specifically optimized for instruction following as our baselines, including IR-1.5B, Conifer-7B-DPO, Crab-7B-DPO, SPAR-8B-DPO, and VERIF-8B. Additionally, we include strong general-purpose models as baselines. We also compare different methods, including SFT, Self-Rewarding, and ProxyReward\footnote{Details about the baselines are provided in Appx.~\ref{appx:rl}.}.




\subsection{In-Domain Performance}

As shown in Tab.~\ref{tab:main}, our method can significantly improve the constraint-following capability of models with different architectures and sizes on in-domain tasks. Across all in-domain benchmarks, our method consistently improves performance. The largest gain is achieved by Qwen2.5-1.5B-Instruct on IFEval (21.6\%).  Our method is effective across models of various sizes, ranging from 1.5B to 14B parameters. For non-reasoning models, Qwen2.5-7B-Instruct-IF shows gains of 5.0, 5.0, and 2.4 on the three benchmarks. Llama-3.1-8B-Instruct-IF achieves 83.0 on IFEval, surpassing SPAR-8B-DPO and approaching VERIF-8B.

For reasoning models distilled from R1, our method also demonstrates substantial improvements. Distill-Qwen-1.5B-IF achieves gains of 16.5, 3.0, and 3.7 on IFEval, CFBench, and FollowBench, respectively. Distill-Qwen-14B-IF achieves the highest performance among distilled models, with gains of 8.8, 9.0, and 6.4. R1-0528-Qwen3-8B-IF reaches 87.1 on IFEval, matching VERIF-8B and outperforming all other baselines. Compared to instruction-following baselines, our method enables trained models to achieve competitive or superior performance. For instance, Qwen2.5-7B-Instruct-IF outperforms Conifer-7B-DPO, and Crab-7B-DPO on IFEval. 

As shown in Tab.~\ref{tab:comparison_sft}, across both Distill-Qwen-7B and Qwen2.5-7B-Instruct, our method consistently achieves the strongest gains in instruction-following performance and constraint adherence. For Distill-Qwen-7B, it improves performance by up to 10.0 and increases CFR by nearly 10\% across benchmarks, substantially outperforming ProxyReward and Self-Rewarding. For Qwen2.5-7B-Instruct, it yields consistent improvements of around 5.0, together with clear CFR gains, while competing methods provide smaller or inconsistent benefits. In contrast, SFT offers limited improvements and even degrades performance on some benchmarks. Overall, these results demonstrate that our method more effectively enhances both task performance and reliable constraint following across model types.




\definecolor{steelblue}{RGB}{70,130,180}
\definecolor{lightgray}{gray}{0.93}

\begin{table*}[t]
\centering
\small
\renewcommand{\arraystretch}{0.9}
\resizebox{\textwidth}{!}{
\begin{tabular}{llcccccc}
\toprule
\multirow{2}{*}{\textbf{Model}} & \multirow{2}{*}{\textbf{Method}} & 
\multicolumn{2}{c}{\textbf{IFEval}} & 
\multicolumn{2}{c}{\textbf{CFBench}} & 
\multicolumn{2}{c}{\textbf{FollowBench}} \\
\cmidrule(lr){3-4} \cmidrule(lr){5-6} \cmidrule(lr){7-8}
& & \textbf{Pr.(L)} & \textbf{CFR} & \textbf{ISR} & \textbf{CFR} & \textbf{HSR} & \textbf{CFR} \\
\midrule
\multirow{5}{*}{Distill-Qwen-7B}
& Base & 61.7 & 70.74\% & 36.0 & 73.14\% & 41.7 & 63.99\% \\
& SFT & 62.7 & 72.30\% & 35.0 & 72.50\% & 39.2 & 58.93\% \\
& Self-Rewarding & 65.1 & 74.10\% & 39.0 & 75.75\% & 44.1 & 66.23\% \\
& ProxyReward & 68.4 & 77.34\% & 40.0 & 76.26\% & 45.4 & 67.91\% \\
& \cellcolor{steelblue!25}Our Method-IF & \cellcolor{steelblue!25}\textbf{71.7} & \cellcolor{steelblue!25}\textbf{80.46\%} & \cellcolor{steelblue!25}\textbf{42.0} & \cellcolor{steelblue!25}\textbf{78.80\%} & \cellcolor{steelblue!25}\textbf{49.1} & \cellcolor{steelblue!25}\textbf{70.67\%} \\
\midrule
\multirow{5}{*}{Qwen2.5-7B-Instruct}
& Base & 73.9 & 81.29\% & 47.0 & 80.02\% & 55.1 & 75.35\% \\
& SFT & 74.5 & 82.13\% & 43.0 & 77.75\% & 54.0 & 74.91\% \\
& Self-Rewarding & 75.2 & 82.25\% & 48.0 & 80.13\% & 55.4 & 76.40\% \\
& ProxyReward & 75.4 & 82.34\% & 49.0 & 80.62\% & 56.2 & 76.68\% \\
& \cellcolor{steelblue!25}Our Method-IF & \cellcolor{steelblue!25}\textbf{78.9} & \cellcolor{steelblue!25}\textbf{85.37\%} & \cellcolor{steelblue!25}\textbf{52.0} & \cellcolor{steelblue!25}\textbf{82.53\%} & \cellcolor{steelblue!25}\textbf{57.5} & \cellcolor{steelblue!25}\textbf{77.47\%} \\
\bottomrule
\end{tabular}
}
\caption{Comparison of different baseline methods. CFR denotes Constraint Following Rate.}
\label{tab:comparison_sft}
\end{table*}

\begin{figure}[t] 
    \centering
            \includegraphics[width=0.3\textwidth]{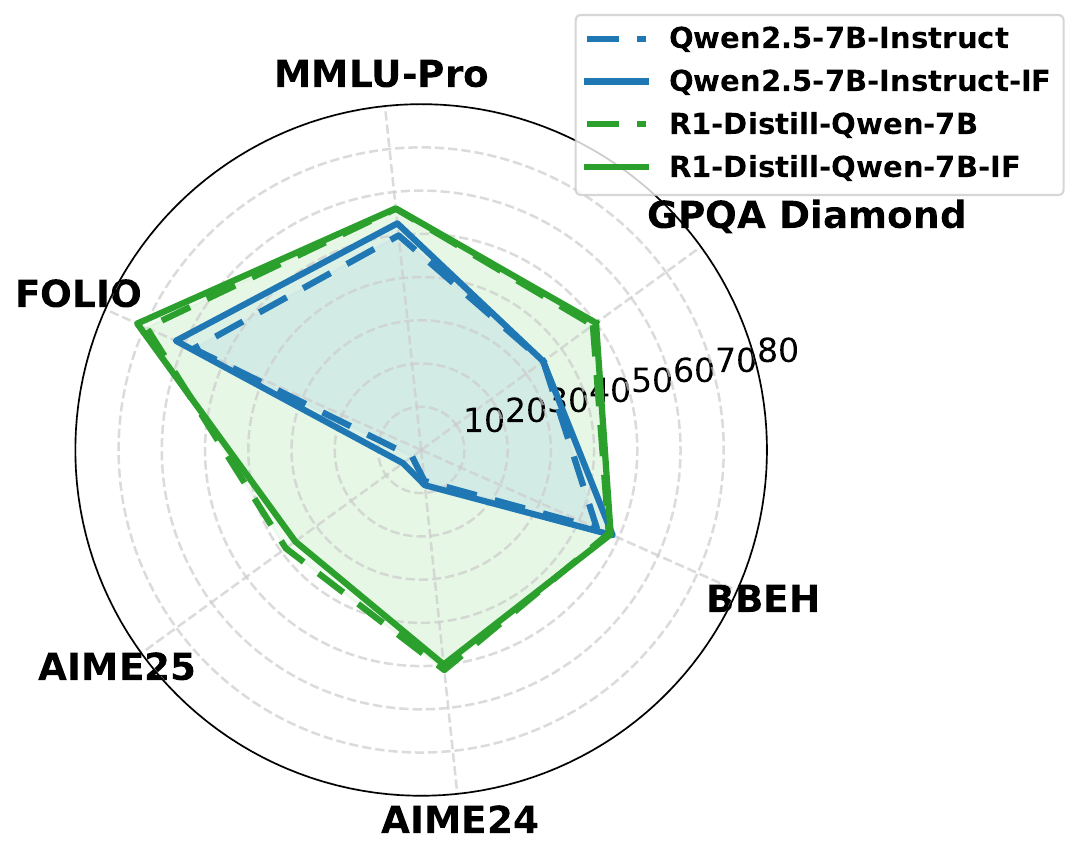}
    \caption{Performance on general benchmarks.}
    \label{fig:com11}
\end{figure}



\definecolor{steelblue}{RGB}{70,130,180}

\begin{table*}[t]
\centering
\resizebox{1.0\textwidth}{!}{
\begin{tabular}{lccccc}
\toprule
\multirow{2}{*}{Method} 
& \multicolumn{1}{c}{IFEval} & \multicolumn{1}{c}{CFBench} 
& \multicolumn{1}{c}{FollowBench} & \multicolumn{1}{c}{ComplexBench} & \multicolumn{1}{c}{Collie}  \\ 
\cline{2-6}
& Pr. (L) & ISR & HSR & Overall & Avg.  \\ 
\midrule

\rowcolors{2}{gray!10}{white}
R1-Distill-Qwen-7B-IF & 71.7 & 42.0 & 49.1 & 60.7 & 27.0 \\ 

\rowcolor{steelblue!25}
\multicolumn{6}{c}{\textit{Ablation Study }} \\ 
\rowcolors{2}{gray!10}{white}
w/o rule\_based reward & -4.2 & -2.0 & -2.4 & -1.7  & -6.9  \\ 
w/o incremental constraint curriculum & -4.0 & -2.0 & -3.6 & -3.0 & -1.9 \\ 

\bottomrule
\end{tabular}
}
\caption{Ablation study results on reward modeling and incremental constraint curriculum.}
\label{tab:abla}
\end{table*}

\begin{table*}[t]
\centering
\resizebox{1.0\textwidth}{!}{%
\begin{tabular}{lccccccccccc}
\toprule
\multirow{2}{*}{\textbf{Reward Model}} & 
\multicolumn{4}{c}{\textbf{4 constraint}} & 
\multicolumn{4}{c}{\textbf{5 constraint}} & \multicolumn{3}{c}{\textbf{Evaluation on Trained Policy Model}}\\
\cmidrule(lr){2-5} \cmidrule(lr){6-9} \cmidrule(lr){10-12}
&
\textbf{KT$\uparrow$}   & \textbf{PC$\uparrow$}   & \textbf{Time$\downarrow$} & \textbf{Speedup$\uparrow$} &
\textbf{KT$\uparrow$}   & \textbf{PC$\uparrow$}   & \textbf{Time$\downarrow$} & \textbf{Speedup$\uparrow$} &\textbf{IFEval}&\textbf{CFBench}&\textbf{FollowBench} \\ 
\midrule
QwQ-32B(Const.-Level)                      & 78.0 & 89.7 & 25.3s & $\times 1.0$   & 74.8 & 89.0 & 37.1s & $\times 1.0$ &72.5&46.0&51.1\\
QwQ-32B(Inst.-Level)                      & 76.7 & 88.7 & 23.0s & $\times 1.1$   & 73.2 & 87.2 & 35.7s & $\times 1.1$ &69.9&45.0&45.8\\
\hdashline
IF-Verifier-7B               & 58.7 & 80.7 & 5.2s  & $\times 4.9$   & \textbf{61.2} & 82.0 & 7.4s  & $\times 5.0$ &69.1&41.0&41.6 \\ 
BT-Training-7B            & 59.3 & 81.7 & 0.3s  & $\times 84.3$  & 48.8 & 78.8 & 0.4s  & $\times 92.8$&65.6& 40.0&39.4 \\ 
Our-RM-7B          & \textbf{62.7} & \textbf{83.3} & \textbf{0.2s}  & $\times 126.5$ & \textbf{61.2} & \textbf{83.4} & \textbf{0.3s}  & $\times 123.7$ &\textbf{71.7}&\textbf{42.0}&\textbf{49.1}  \\ 
\bottomrule
\end{tabular}%
}
\caption{Comparison of reward models on human-alignment (Kendall's Tau, Position Consistency), inference speed under varying constraints, and RL training performance on Distill-Qwen-7B. Const.-Level and Inst.-Level refer to constraint-level and instruction-level evaluations, respectively.}
\label{tab:rm}
\end{table*}

\begin{figure}[t] 
    \centering
            \includegraphics[width=0.5\textwidth]{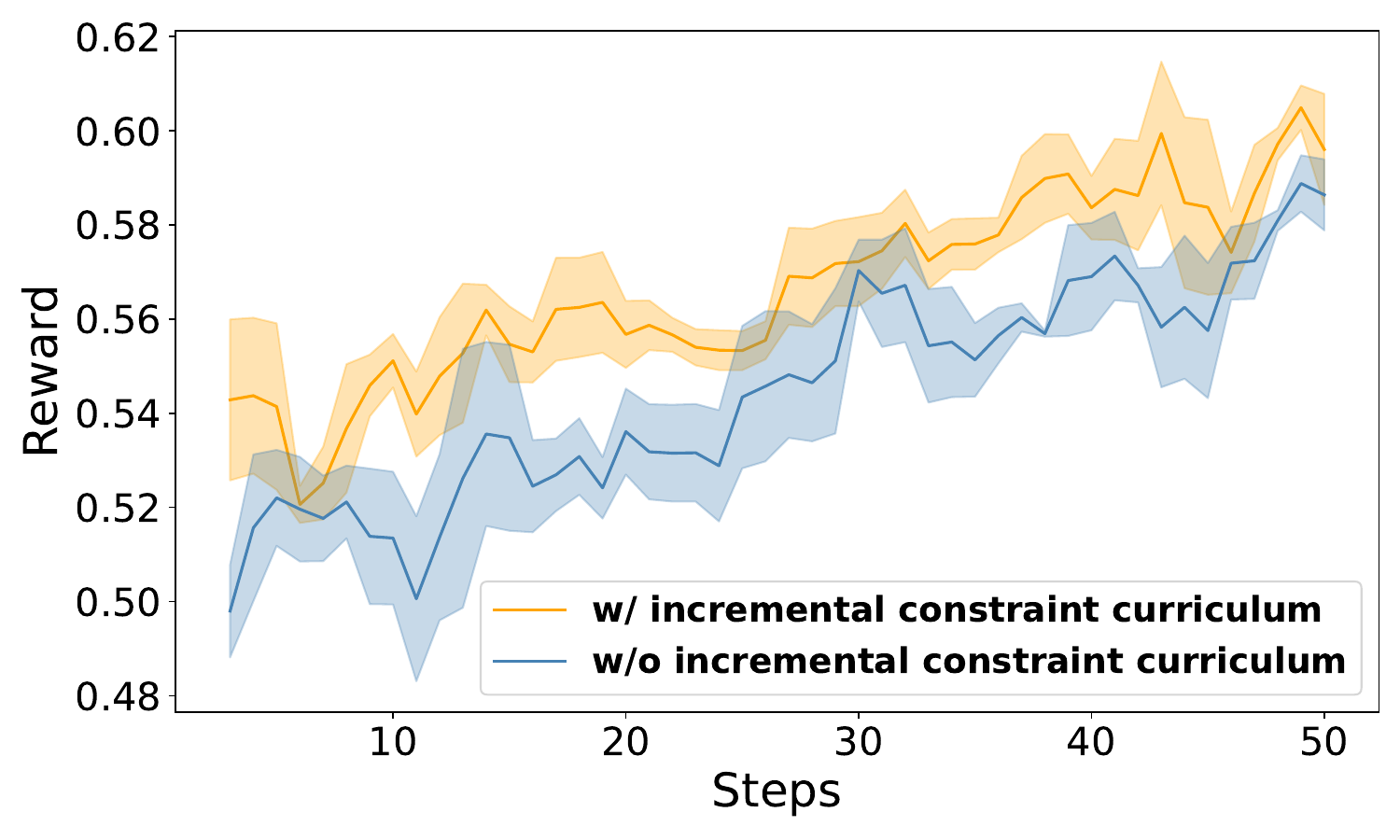}
    \caption{Comparison of reward density.}
    \label{fig:com121}
\end{figure}
\begin{figure*}[t] 
    \centering
            \includegraphics[width=1.0\textwidth]{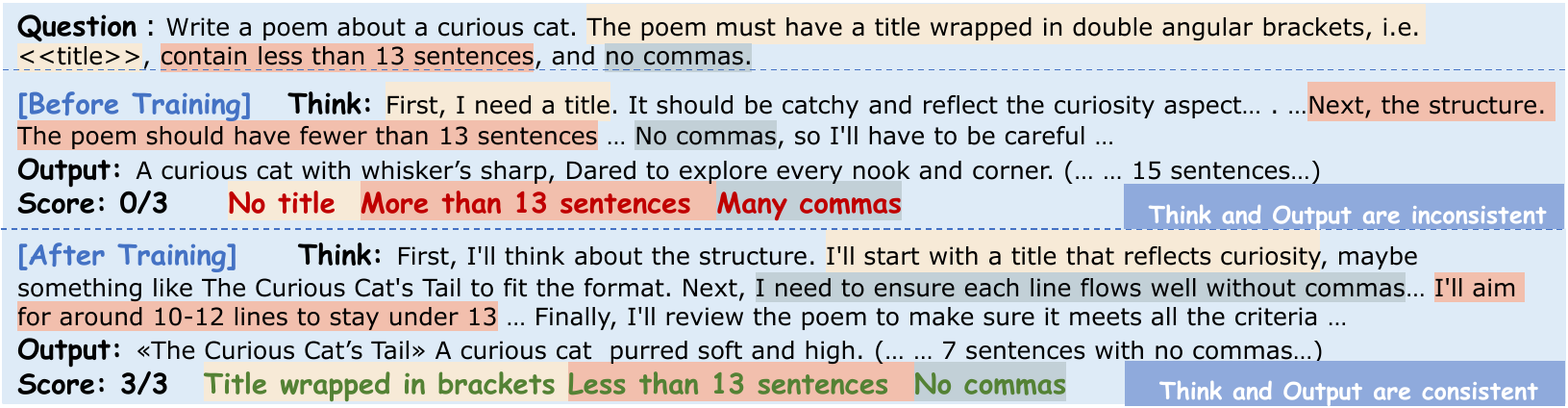}
    \caption{Qualitative analysis of  consistency between  model's thinking and outputs before and after training.}
    \label{fig:case}
\end{figure*}

\subsection{Generalizability}
\paragraph{\textbf{Out-of-Domain Generalization.}}

We select benchmarks containing constraints  different from those in the training data to evaluate the model’s out-of-domain generalization ability, including writing tasks, agent tasks, and multi-turn dialogues. As shown in Tab.~\ref{tab:main}, our method improves the model's instruction-following ability on out-of-domain tasks. Specifically, after training, R1-0528-Qwen3-8B achieves performance gains of 6.1 and 7.3 on AgentIF and MultiChallenge, respectively.


\paragraph{\textbf{General Abilities.}}
Besides instruction-following ability, we also evaluate the model's general ability\footnote{Details about the benchmarks are in Appx.~\ref{appx:bec}.}. As shown in Fig.~\ref{fig:com11}, we evaluate AIME using Avg@30 method, which refers to the average score across 30 sampled responses per question. The max output tokens is set to 32k. The results show that our method can maintain the model's general abilities. On certain benchmarks, the model’s performance declines because instruction-following ability and general reasoning ability are two distinct capabilities — fine-tuning to enhance one often comes at the expense of the other.

\subsection{Ablation Studies}

\paragraph{\textbf{Incremental Constraint Curriculum.}}
To validate the effectiveness of incremental constraint curriculum, we conduct ablation studies by removing this component and directly training on multi-constraint instructions without decomposition. As shown in Tab.~\ref{tab:abla}, removing the incremental constraint curriculum leads to performance drops. As illustrated in Fig.~\ref{fig:com121}, the model trained w/o incremental constraint curriculum receives sparser rewards, which hinders its ability to learn to follow constraints.

\paragraph{\textbf{Rule\_based Reward Modeling.}}
We conduct ablation studies for reward modeling under the following settings: \textbf{w/o rule\_based reward}, which uses only the reward model to provide rewards. As shown in Tab.~\ref{tab:abla}, this leads to performance drops on these benchmarks. This indicates explicit rule-based verification plays a crucial role in reliably validating hard constraints.

\paragraph{\textbf{Comparison of different reward modeling methods.}} We manually annotated 50 groups in the 5-constraint setting and 50 groups in the 4-constraint setting. In each group, an instruction is paired with a fixed number of constraints, along with candidate responses that satisfy from 1 up to all of the given constraints. We then compare the correlation between human-labeled rankings and the rankings produced by three modeling methods: (a) LLM-as-a-judge, (b) reward model trained with Bradley-Terry (BT) Loss, and (c) our reward model. We use two metrics to measure the consistency with human preferences: \textbf{Kendall Tau coefficient (KT)} measures rank correlation between model rankings and human rankings, and \textbf{Position Consistency (PC)} measures the consistency of relative positions. As shown in Tab.~\ref{tab:rm}, our reward model achieves the best human alignment among 7B models, achieves the fastest inference speed , and leads to the best RL training performance among 7B models . While QwQ-32B achieves slightly higher human alignment scores, its extremely slow inference speed.

\begin{figure}[t]
    \centering
    \includegraphics[width=0.5\textwidth]{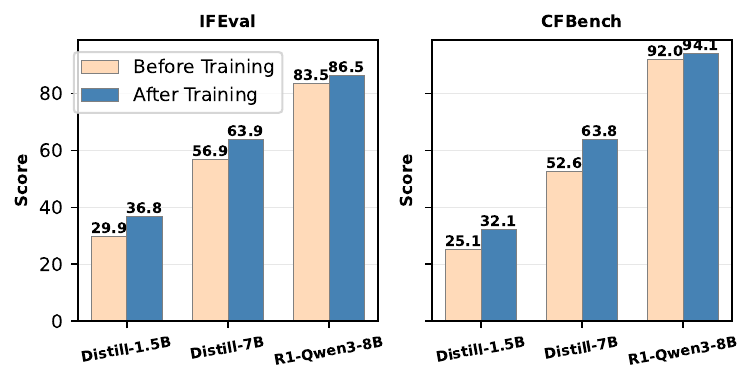}
    \caption{Quantitative analysis of consistency scores before and after model training. }
    \label{fig:analyze11}
\end{figure}


\begin{figure*}[t]
        \centering\includegraphics[width=\textwidth]{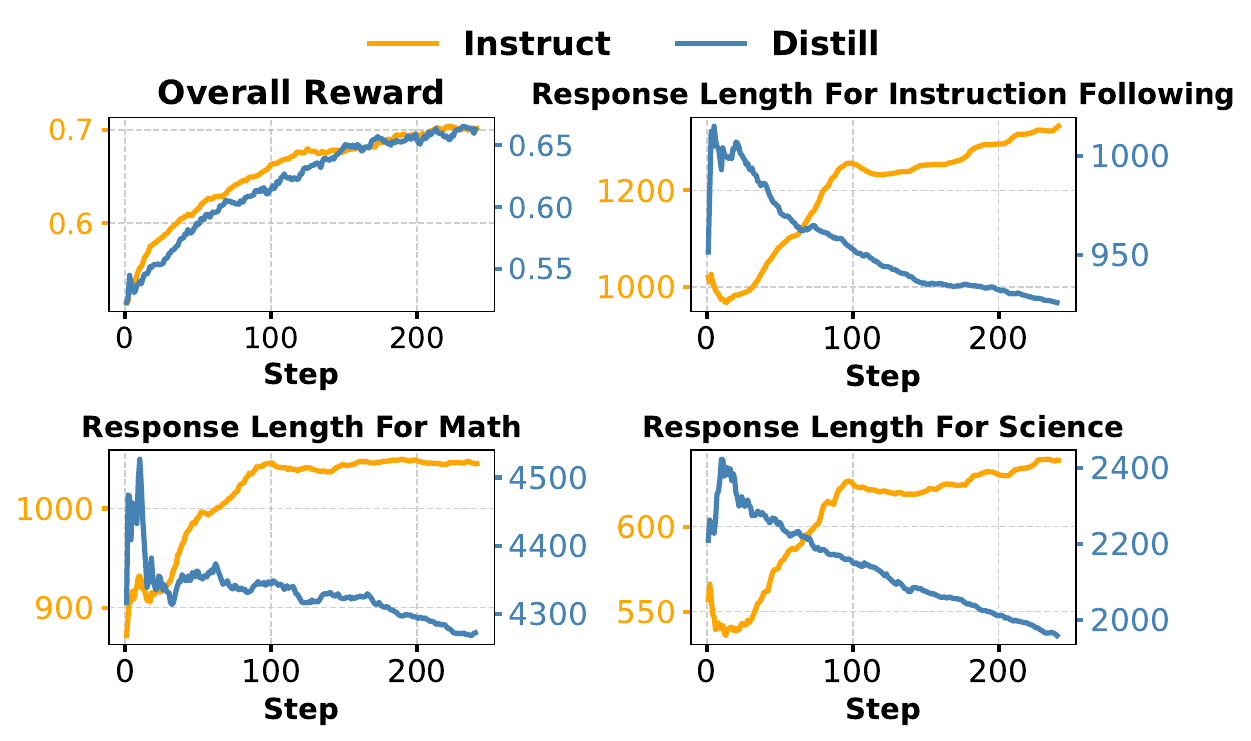}
        \caption{The reward and response length dynamics of Qwen2.5-7B-Instruct and R1-Distill-Qwen-7B.}
        \label{fig:com1}

\end{figure*}

\subsection{Analysis}


\subsubsection{Consistency Analysis}
We analyze the reasons for performance improvement by examining the consistency between the model's thinking process and its generated outputs. From the \textbf{qualitative} perspective, as shown in Fig.~\ref{fig:case}, we present a concrete example of a poem generation task with three constraints: the poem must have a title wrapped in double angular brackets, contain less than 13 sentences, and no commas.

Before training, the model's thinking process demonstrates awareness of all constraints. In its internal reasoning, it acknowledges the need for a title, plans to keep the poem under 13 sentences, and notes the requirement to avoid commas. However, the actual output diverges significantly from this thinking: the generated poem contains no title, exceeds 13 sentences (15 sentences), and includes many commas, resulting in a score of 0/3. This inconsistency between thinking and output indicates that while the model can recognize constraints during reasoning, it fails to translate this understanding into its generated responses. After training, the model exhibits remarkable improvement in consistency. Its thinking process becomes more strategic and structured, explicitly planning the title format, sentence count, and comma avoidance. More importantly, the generated output now perfectly aligns with this thinking: the poem includes a title wrapped in brackets (``«The Curious Cat's Tail»''), contains only 7 sentences (well under the 13-sentence limit), and has no commas, achieving a perfect score of 3/3. The consistency between thinking and output demonstrates that training enables the model to bridge the gap between constraint understanding and constraint execution.

From the \textbf{quantitative} perspective, as shown in Fig.~\ref{fig:analyze11}, we use GPT-4.1 to assess the consistency between the model's reasoning process and its responses on IFEval and CFBench. After training, our method significantly enhances the consistency between the model's thinking and its responses across a wide range of tasks. This improved consistency explains both the effectiveness of our method in improving constraint-following performance and its strong generalization to diverse out-of-domain benchmarks, as the ability to maintain alignment between reasoning and generation is fundamental to reliable instruction following.

\subsubsection{Training Dynamics}

As shown in Fig.~\ref{fig:com1}, we compare the training dynamics of Qwen2.5-7B-Instruct and R1-Distill-Qwen-7B. For both models, rewards increase steadily from around 0.55 to approximately 0.7 by step 200, where they converge and plateau. Despite similar reward gains, the two models exhibit distinct response-length dynamics across tasks. For \textbf{instruction-following tasks}, the instruct model shows a monotonic increase in response length, indicating that RL progressively induces more explicit reasoning. In contrast, the distilled model briefly expands early in training and then steadily contracts to below 950 tokens, reflecting a shift toward concise responses.

For \textbf{math tasks}, the instruct model again gradually increases response length, consistent with emerging mathematical reasoning. The distilled model starts from a much higher baseline, peaks early, and then declines, suggesting improved efficiency over already strong reasoning abilities. A similar trend is observed for \textbf{science tasks}: the instruct model shows modest growth, while the distilled model exhibits an early peak followed by a sustained reduction from a high initial baseline. Overall, while both models reach comparable reward levels, their learning trajectories differ fundamentally.

\section{Conclusion}
We propose a self-supervised RL framework that enhances instruction following without external supervision signals.  Our approach features: (1) instruction decomposition for dense training signals, (2) self-supervised reward modeling using pseudo-labels generated directly from the instructions themselves, and (3) constraint-wise binary classification for efficiency.
Experiments show significant improvements in instruction following while preserving general capabilities. Moreover, our approach generalizes well, yielding strong gains across 3 in-domain and 5 out-of-domain benchmarks, including challenging agentic and multi-turn instruction-following tasks.

\section*{Limitations}

Due to computational resource limitations, we have not validated our method on larger-scale models (e.g., 72B parameters), though our experiments on smaller models provide strong evidence of the method's effectiveness and scalability potential.
Additionally, as our primary focus centers on reward modeling design, the construction of multi-constraint datasets remains relatively limited in diversity, in terms of task types and constraint types. In the real world, user instructions may include more complex task scopes and constraints.

\section*{Ethical Considerations}

We discuss potential ethical concerns as follows: Ranking on the sampled data from the reward model training set was conducted by three annotators recruited by our institution. The construction team remains anonymous to
the authors. We ensure that the privacy rights of all
annotators are respected throughout the annotation
process. All annotators are compensated above
the local minimum wage and consent to use the data for research purposes. The annotation details are
shown in Appx.~\ref{appx:es}.






\bibliography{custom}

\appendix
\clearpage
\newpage
\section{Appendix}
\subsection{Dataset Analysis}

\begin{figure*}[t] 
    \centering
            \includegraphics[width=1\textwidth]{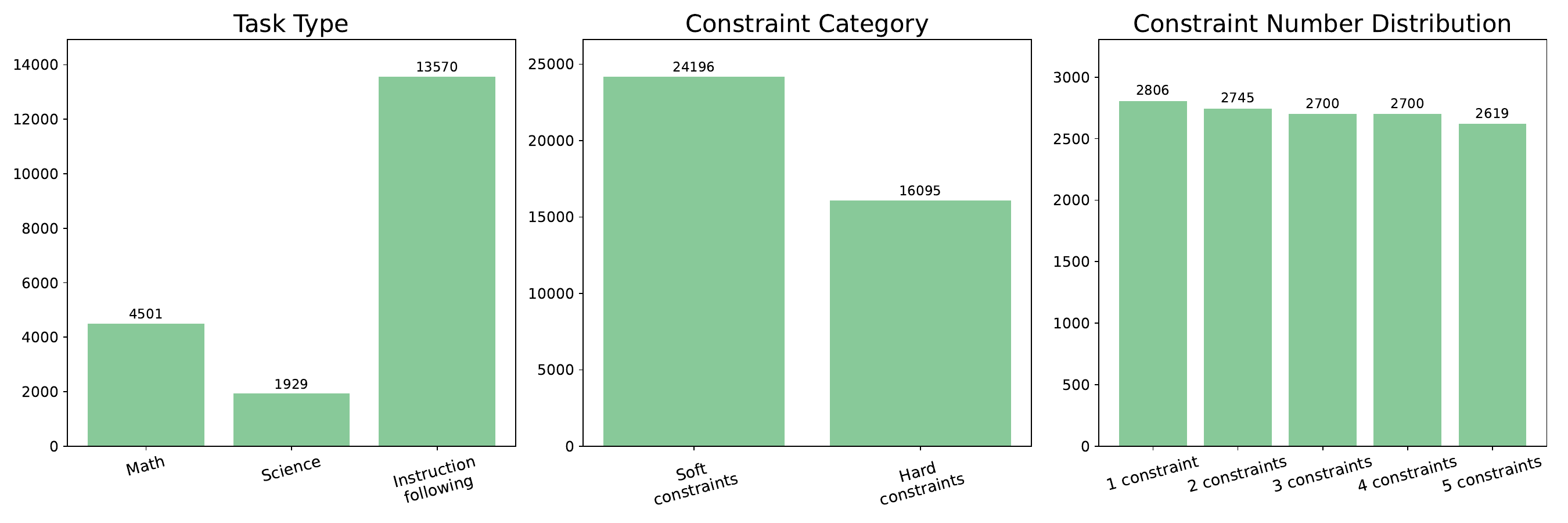}
    \caption{Sample Distribution across Task Types, Constraint Categories, and Number of Constraints}
    \label{fig:dataset}
\end{figure*}

\subsubsection{Constraint Distribution}
Our dataset includes instruction-following, mathematical, and scientific tasks.  For instruction-following tasks, the constraints added in the instructions can be classified into soft constraints and hard constraints. Each instruction contains 1 to 5 constraints. As shown in Fig.~\ref{fig:dataset}, we present statistics on the amount of data across different domains, the number of soft and hard constraints, and the distribution of instructions with varying numbers of constraints. As shown in Tab.~\ref{tab:app_soft} and Tab.~\ref{tab:app_hard}, we present examples of soft and hard constraints in the dataset. 

\subsubsection{Constraint Types}
\label{appx:ct}
As shown in Tab.~\ref{tab:constraint-types}, our dataset includes various types of constraints, covering multiple domains and tasks. For each seed instruction, we use the prompt shown in Tab.~\ref{tab:prompt-raw} to construct constraints, resulting in multi-constraint instructions.

To ensure constraint compatibility when synthetically adding multiple constraints to a seed instruction, we employ two mechanisms: (1) \textbf{Constraint Taxonomy:} We define the taxonomy for soft and hard constraints ahead of time, which helps avoid obvious conflicts while constructing the dataset within this taxonomy. (2) \textbf{Flexible Constraint Selection:} The constraint construction prompt explicitly states that "[Constraint References] are just suggestions. You may choose one or more constraints from the list or propose new ones if needed." This flexibility keeps the model from being forced to compose incompatible constraints and further reduces the chance of conflicts.

\begin{table*}
\resizebox{\linewidth}{!}{
\begin{tcolorbox}
\small

\textit{\color{gray}{/* Seed instruction */}} \\
Assign a category to each text snippet identifying the main emotion expressed. Categories may include ``joy,'' ``anger,'' ``sadness,'' or ``fear.''  Sentence: NAME\_2 was overwhelmed with tears of happiness as she accepted the award on stage.

\textit{\color{gray}{/* Constraints */}} \\
Must include the word ``exuberant'' in at least one category description.

Incorporate a scenario where a character is experiencing mixed emotions.

Limit the response to a maximum of 25 words.

Write the response in the style of a psychological evaluation report.

Tailor the answer for an audience of high school psychology students.

\end{tcolorbox}
}
\caption{Examples of instructions with soft constraints.}
\label{tab:app_soft}
\end{table*}
\begin{table*}
\resizebox{\linewidth}{!}{
\begin{tcolorbox}
\small

\textit{\color{gray}{/* Seed instruction */}} \\
Given a question, generate a paraphrase of that question without changing the meaning of it. Your answer should reword the given sentence, but not add information to it or remove information from it. The answer to your question should be the same as the answer to the original question. Input: Question: who does the voice of carl in phineas and ferb?

\textit{\color{gray}{/* Constraints */}} \\
Answer with the letter \texttt{"i"} appearing at least 4 times. 

Highlight at least 2 sections of your response in markdown such as \texttt{*highlighted section*}.

Make sure your reply is in English and all capital letters. 

Put your entire response inside double quotation marks.

\end{tcolorbox}
}
\caption{Examples of instructions with hard constraints.}
\label{tab:app_hard}
\end{table*}
\begin{table*}
\centering
\renewcommand{\arraystretch}{1.5}
\small
\begin{tabular}{|p{3.5cm}|p{4.8cm}|p{5.5cm}|}
\hline
\textbf{Constraint Type} & \textbf{Definition} & \textbf{Example} \\
\hline
Lexical content constraint & Must include specific terms or symbols with precise placement. & "...must include the word 'beautiful.'" \\
\hline
Element constraint & Include specific entities or scenarios. & "...highlights the Great Wall." \\
\hline
Semantic constraint & Focus on themes, tone, or stance. & "Write a poem about London." \\
\hline
Word Count & Limit the number of words. & "A 50-word poem." \\
\hline
Sentence Count & Limit the number of sentences. & "...three sentences." \\
\hline
Paragraph Count & Limit the number of paragraphs. & "...divided into 3 sections." \\
\hline
Document Count & Limit the number of documents. & "...list 3 articles." \\
\hline
Tone and emotion & Conform to specific emotional tone. & "Write a letter in an angry and sarcastic tone." \\
\hline
Form and style & Use specified stylistic form and perception. & "Write a passage in an encyclopedic style." \\
\hline
Audience-specific & Tailored to a specific audience group. & "Write a poem for a 6-year-old." \\
\hline
Authorial style & Emulate specific authors’ styles. & "Write a passage in the style of Shakespeare." \\
\hline
Fundamental format & Follow standard formats like JSON, HTML, etc. & "Output in JSON format." \\
\hline
Bespoke format & Use custom formatting protocols. & "Bold the main idea and output in unordered list." \\
\hline
Specialized format & Tailored for specific applications or domains. & "Convert to electronic medical record format." \\
\hline
Pragmatic constraint & Adapt to context like dialects or language policy. & "Output in English, classical Chinese, etc." \\
\hline
Syntactic constraint & Follow specific phrase and clause structures. & "Use imperatives with nouns and verb phrases." \\
\hline
Morphological constraint & Control over affixes, roots, and word formation. & "Output all content in lowercase English." \\
\hline
Phonological constraint & Focus on sounds, tone, and intonation. & "Single-syllable tongue twisters." \\
\hline
Role-based constraint & Respond with specific role identity. & "You are Confucius, how do you decide?" \\
\hline
Task-specific constraint & Address a defined situational task. & "Work from home, how to report?" \\
\hline
Complex context constraint & Involve multi-faceted and nested reasoning. & "On the left, 10 total, what to do?" \\
\hline
Example constraint & Conform to patterns from example pairs. & "input:x..., output:{...}; input:y..., output?" \\
\hline
Inverse constraint & Narrow response space via exclusions. & "No responses about political topics." \\
\hline
Contradictory constraint & Combine requirements that are hard to satisfy simultaneously. & "A five-character quotation, 1000 words." \\
\hline
Rule constraint & Follow symbolic or logical operation rules. & "Each answer adds 1+1=3, then 2+2=5." \\
\hline
\end{tabular}
\caption{Constraint types in our training data.}
\label{tab:constraint-types}
\end{table*}

\begin{table*}
\centering
\renewcommand{\arraystretch}{1.2}
\small
\begin{tabular}{|p{15cm}|}
\hline
[Task Description] \\
1. I currently have a seed question, but the seed questions are relatively simple. To make the instructions more complex, I want you to identify and return five atomic constraints that can be added to the seed question. \\
2. I will provide [Seed Question] and [Constraint References], and you can use these references to propose five constraints that would increase the difficulty of the seed question. \\
3. [Constraint References] are just suggestions. You may choose one or more constraints from the list or propose new ones if needed. \\
4. Do not modify or rewrite the seed question. Your task is only to generate new constraints that can be added to it. \\
5. Return the added constraints in the following JSON format: \\
\quad json \\
\quad \{ \\
\quad\quad "c1": "<first constraint>", \\
\quad\quad "c2": "<second constraint>", \\
\quad\quad "c3": "<third constraint>", \\
\quad\quad "c4": "<fourth constraint>", \\
\quad\quad "c5": "<fifth constraint>" \\
\quad \} \\
6. Do not return anything else. No explanation, no reformulated question, no analysis---only the JSON structure. \\
\\

[Constraint References] \\
1. Lexical content constraint : \{Definition\} \{Example\} \\
2. Word Count : \{Definition\} \{Example\} \\
$\ldots$ \\
25. Rule Constraint : \{Definition\} \{Example\} \\
\\

[Seed Question] \\
\{raw\_question\} \\
\hline

\end{tabular}
\caption{Prompt for generating constraints.}
\label{tab:prompt-raw}
\end{table*}

\begin{table*}[t]
\centering
\begin{tabular}{p{0.22\linewidth} p{0.70\linewidth}}
\hline
\textbf{Item} & \textbf{Description} \\
\hline
Task Overview & In this study, you will evaluate 50 groups of responses. Each group contains one instruction with five explicit constraints and five candidate responses. You will rank the five responses according to how well they satisfy the constraints in the instruction. \\

What to Do & (1) You will read the instruction carefully and identify the five constraints. (2) You will read all five candidate responses. (3) You will judge how many constraints each response satisfies. (4) You will rank the responses from best to worst based on overall constraint satisfaction. A response that satisfies more constraints should be ranked higher than one that satisfies fewer constraints. \\

Annotation Criteria & You should focus only on the constraints explicitly stated in the instruction, without adding extra assumptions or personal preferences. If two responses appear very similar, you should rank higher the one that more clearly and completely satisfies the stated constraints. You should base your judgment only on the text shown in the task. \\

Output Format & For each group, you will provide a ranking of the five responses from best to worst. For example: \texttt{Response B > Response E > Response A > Response D > Response C}. \\

Interface  & You will see one instruction at the top of the page and the five candidate responses listed below it, each with a response ID. A ranking input field or selection interface will be provided for submitting the final ordered list. \\
\hline
\end{tabular}
\caption{Instructions given to participants for the preference ranking task.}
\label{tab:participant_instructions}
\end{table*}

\begin{table*}
\resizebox{\linewidth}{!}{
\begin{tcolorbox}
\small

\textit{\color{gray}{/* Positive Example */}} \\
\textbf{Question:} Introduce Time Constraints: Impose a time limit for the role-play to unfold or for certain actions to be completed. \\
\textbf{Answer:} Characters identified in the story: 1. Kathy 2. Sue. The main conflict between the characters is the tension and argument that arises due to Kathy spreading rumors about Sue and Sue discovering and reacting to these rumors. To introduce a time constraint for the role-play, we can say: The events in the story unfold over the course of a week. Kathy starts the rumors on Monday, Sue finds out by Wednesday, they have the argument on Thursday, and by the end of the week, Sue stops speaking to Kathy. This time constraint adds a sense of urgency and progression to the narrative. \\
\textbf{Label:} 1

\textit{\color{gray}{/* Negative Example */}} \\
\textbf{Question:} Introduce Time Constraints: Impose a time limit for the role-play to unfold or for certain actions to be completed. \\
\textbf{Answer:} The characters available in the given story are Kathy and Sue. The main conflict between the characters is the tension and argument that arises due to Kathy's rumors about Sue, which leads to Sue finding out and subsequently never talking to Kathy again. \\
\textbf{Label:} 0

\end{tcolorbox}
}
\caption{Examples of reward model training data.}
\label{tab:app_rm}
\end{table*}

\begin{table}[ht]
\centering
\resizebox{0.4\textwidth}{!}{
\begin{tabular}{lc}
\hline
\textbf{Model Name} & \textbf{Test Accuracy} \\
\hline
Original Reward Model & 85.3\% \\
Manually Annotated Model & 86.1\% \\
\hline
\end{tabular}
}
\caption{Comparison of the performance of reward models trained on pseudo-labeled data and human-labeled data.}
\label{tab:reward-model-comparison}
\end{table}

\subsubsection{Evaluation Setup}
\label{appx:es}
We sampled 50 groups of preference data from the reward model training set. Each group consisted of one instruction with five constraints and five corresponding responses. The five responses were generated under progressive constraint settings: the first response was generated using only the first constraint, the second using the first two constraints, and so on, until the fifth response was generated using all five constraints. We recruited three students in computer science as annotators to manually annotate these 50 groups. The annotators were paid  above the local minimum wage and consented to use the data for research purposes
covered in our paper. The instructions given to participants are shown in Tab.~\ref{tab:participant_instructions}. When annotations conflict, we adopt a majority-vote approach to decide the final outcome. We then compared the correlation between the human-labeled ranking and the reference ranking induced by the progressive constraint settings.

\subsection{Reward Model Training}
\label{appx:rmt}
We perform full fine-tuning on three models, Qwen2.5-1.5B-Instruct, Qwen2.5-7B-Instruct, and Llama-3.1-8B-Instruct, for a binary classification task aimed at determining whether a response satisfies a given constraint. Each training example consists of a prompt that concatenates the response and its associated constraint. Fine-tuning is conducted using the HuggingFace Trainer framework with FP16 precision enabled and Deepspeed optimization (Stage 2 ZeRO) configured via a JSON file specifying automatic batch size and gradient accumulation steps, as well as support for the adamw\_torch optimizer. Training uses a batch size of 1, a learning rate of 5e-6, and runs for 3 epochs. Accuracy is employed as the evaluation metric. The Qwen2.5-1.5B reward model provides reward signals for reinforcement learning (RL) training of 1.5B models with the same backbone, the Llama-3.1-8B reward model provides reward signals for RL training of Llama-3.1-8B-Instruct, while the Qwen2.5-7B reward model is used for the RL training of other Qwen-based models of the same size or larger. All training is performed on 8 NVIDIA A100 80GB GPUs. As shown in Tab.~\ref{tab:app_rm}, we present examples of the reward model training data. For the reward model training, we totally use 14,058 samples. 

As shown in Tab.~\ref{tab:reward-model-comparison}, for the reward model training data, we performed manual annotation and trained the reward model under the same setting, with 10\% of the data split off as the test set. The accuracy of the reward model trained with manually labeled data showed no significant improvement than our original reward model on the test set. Although the reward model's training data may contain noisy labels, this does not significantly compromise its overall performance. This demonstrates the advantage of our pseudo-labeled reward model training data construction: it achieves performance comparable to human-annotated baselines while eliminating the need for manual labeling.

\subsubsection{Self-Supervised Training Framework}
\label{appx:self-supervision}
Our "self-supervised" approach utilizes label-free instructions as input. The self-supervision refers to the supervision of the model's responses and labels, not the supervision of the instructions themselves. As defined above: "Our self-supervision supervises the outputs and labels, not the instructions." Furthermore, as seen in Tab.~\ref{tab:comparison_sft}, performing SFT directly using the generated data does not significantly improve model performance. This suggests that the use of GPT-4o for instruction generation provides limited gains on its own; the primary improvement comes from our training methodology.

The performance gains stem from two key components: (1) \textbf{Role of Reward Modeling:} As shown in Tab.~\ref{tab:comparison_sft}, SFT yields minimal performance gains and even degradation on some benchmarks. In contrast, our reward training led to significant improvements. Additionally, based on the main results and the comparison of different reward modeling methods in Tab.~\ref{tab:rm}, our reward modeling component is critical for performance. (2) \textbf{Role of Data Augmentation:} Our constraint curriculum data augmentation method is also instrumental. According to the ablation studies in Tab.~\ref{tab:abla}, training directly on raw data without curriculum augmentation results in a noticeable drop in performance. Both our reward modeling and data augmentation strategies are essential and contribute significantly to the final performance improvements.

\subsubsection{Pseudo-label Quality and Robustness}
\label{appx:pseudo-label-handling}
The core assumption for our reward model training is that "for constraint $c_k$, the response $o_k$ (generated for the instruction with constraint $c_k$) is likely to satisfy it." Our reward model is trained with pseudo labels—essentially a weakly supervised setup—so the framework is designed to tolerate a small amount of noise. The noise level is low: our pseudo labels agree with human preference judgments in over 90\% of cases shown in Tab.~\ref{tab:human_label}. Furthermore, as shown in Tab.~\ref{tab:reward-model-comparison}, the comparison of the performance of reward models trained on pseudo-labeled data versus fully human-labeled data indicates that swapping in fully human-labeled data improves test performance by less than 1\%. This confirms both the quality of our weakly supervised data and the validity of the pseudo-label assumption, even when the number of constraints is high (e.g., four or five).

\subsubsection{Comparison with Large Language Model Judges}
\label{appx:qwq-comparison}
As shown in Tab.~\ref{tab:rm}, we compare our trained reward model with QwQ-32B, a large language model used as a judge. While QwQ-32B demonstrates better correlation with human preferences, its resulting RL training performance on the policy model is inferior to our proposed reward model. Following VERIF, the QwQ-32B baseline uses instruction-level scoring: it outputs 1 only if all constraints are satisfied, otherwise 0. This yields higher Kendall's Tau (KT) and Pearson Correlation (PC) but produces extremely sparse rewards for RL. Our trained reward model works at the constraint level (averaging over individual constraints), delivering denser signals and improving RL performance. It is also about 100× faster than QwQ-32B, achieving both effectiveness and efficiency.

We also evaluate QwQ-32B when used as a constraint-level reward. The results show that constraint-level rewards are denser than instruction-level ones, improving both alignment metrics and RL outcomes. Moreover, our self-supervised 7B reward model matches the effectiveness of a powerful judge like QwQ-32B while being 100× faster, validating its practicality. However, our approach is grounded in self-supervised RL—we do not rely on external large models during training. Using QwQ-32B as an "LLM-as-a-judge" introduces an external supervision signal, which diverges from this goal.

\subsection{RL Training}
\label{appx:rl}
\subsubsection{Implementation Details}
We apply the GRPO training using the VeRL framework. We use a distributed training setup across 3 nodes, each equipped with 8 NVIDIA A100 80GB GPUs, for a total of 24 NVIDIA A100 80GB GPUs. Prompts and responses are truncated to a maximum length of 8192 tokens. The optimizer uses a learning rate of 1e-6 with a weight decay of 1e-2, and no learning rate warm-up. We leverage Fully Sharded Data Parallel (FSDP) with full sharding enabled, rank0 parameter initialization, and parameter offloading to optimize GPU memory usage. The training batches are organized with a global batch size of 96, micro-batches of size 2 for updates, and micro-batches of size 16 for experience generation. Gradient clipping is applied with a max norm of 1.0. Rollouts are performed with a temperature of 1.0 and a group size of 5. Tensor parallelism of size 2 is applied. For Qwen2.5-7B-Instruct, Distill-Qwen-14B, Llama-3.1-8B-Instruct, and Qwen2.5-32B-Instruct, we trained for 52 steps. For Qwen2.5-1.5B-Instruct, R1-0528-Qwen3-8B, Distill-Qwen-1.5B, and Distill-Qwen-7B, we trained for 260 steps.

\subsubsection{Scalability to Larger Models}
\label{appx:scalability}
As shown in Tab.~\ref{tab:main}, we conducted experiments on Distill-Qwen-14B and Qwen2.5-32B-Instruct, demonstrating that our method remains effective at larger scales. The experimental results show consistent improvements across multiple benchmarks. These results demonstrate that our approach scales to larger models while maintaining efficiency gains.

\subsubsection{Baseline Models}
\label{appx:bl}
IR-1.5B~\cite{qin2025incentivizing}: A systematic method to boost LLMs in
 dealing with complex instructions via incentivizing reasoning for test-time compute scaling.
 
Conifer-7B-DPO~\cite{sun2024conifer}: The method first performs SFT training on a constructed dataset arranged from easy to hard, and then performs DPO training using an open-source preference dataset.

Crab-7B-DPO~\cite{qi2024constraint}: the method adopts Llama3-70B-Instruct to back-translate constraints and create a high-quality, complex instruction-response dataset for DPO training.

SPAR-8B-DPO~\cite{cheng2024spar}: SPaR is a self-play framework integrating tree-search self-refinement to yield valid and comparable preference pairs free from distractions. By playing against itself, an LLM employs a tree-search strategy to refine its previous responses with respect to the instruction while minimizing unnecessary variations.

VERIF~\cite{peng2025verif}: VERIF is a verification method that combines rule-based code verification with LLM-based verification from a large reasoning model (RLVR). While both VERIF and our method employ reinforcement learning for instruction following, there are several key differences: (1) \textbf{No reliance on external supervision:} VERIF uses QwQ-32B as an "LLM-as-a-judge" to provide rewards and then distills its outputs into a smaller IF-Verifier-7B reward model. Both settings depend on a strong external model. In contrast, our self-supervised framework trains the reward model without any external supervision signals. (2) \textbf{Denser rewards:} Multi-constraint instructions are hard to satisfy in one shot. VERIF directly uses such complex prompts as the training data. We instead decompose multi-constraint instructions via an Incremental Constraint Curriculum, making learning easier. Our approach yields much denser rewards, which in turn supports stronger RL performance. (3) \textbf{Higher efficiency:} Our reward model runs 30× faster than IF-Verifier-7B and 100× faster than QwQ-32B, providing dramatic speedups without losing fidelity. (4) \textbf{Better Performance:} Under the same base architecture, our model consistently outperforms VERIF, especially on out-of-domain instruction-following benchmarks.

\subsubsection{Baseline Methods}
\label{appx:bm}

SFT is a supervised fine-tuning approach that uses larger, more capable models to generate high-quality response targets for training smaller models. For Distill-Qwen-7B, we used QwQ-32B to generate response targets; for Qwen2.5-7B-Instruct, we used Qwen2.5-32B-Instruct to generate responses. The training is conducted using LLaMAFactory~\cite{zheng2024llamafactory}.

Self-as-Judge is a simple RL approach that uses the model itself as the reward model. This baseline directly uses the model to evaluate its own responses without training a separate reward model, providing a straightforward comparison to demonstrate the effectiveness of our trained reward model.

Self-Rewarding~\cite{yuan2024self} is a method where the language model generates its own training data and rewards to iteratively improve itself over multiple rounds.

ProxyReward~\cite{guo2025general} is a method that pre-generates multiple question-answer (QA) pairs for each meta-question, evaluates response quality based on these pairs, and relies on a strong model for both generation and reward estimation. The approach works by first using a powerful model to generate multiple QA pairs that serve as reference points for evaluating the quality of generated responses. When training a smaller model, ProxyReward assesses the generated responses by comparing them against these pre-generated QA pairs, focusing on metrics such as information comprehensiveness and accuracy. The reward signal is computed based on how well the generated response addresses the questions in these QA pairs.

\subsubsection{Prompts for Baseline Methods}
All baseline methods utilize the same generated dataset as the pseudo-label reward method proposed in this work. We have conducted additional experiments on Qwen2.5-7B-Instruct and Distill-Qwen-7B to compare our method against these baselines. The experimental results demonstrate that our Pseudo-label Reward method outperforms existing reward modeling approaches, significantly enhancing the constraint following rate.

\textbf{1. Self-Rewarding Method}

The Self-Rewarding method employs the prompt described in Fig. 2 of the source paper~\cite{yuan2024self} (``LLM-as-a-Judge prompt for our LLM to act as a reward model and provide self-rewards for its own model generations''). The specific prompt content is shown in Tab.~\ref{tab:prompt-self-rewarding}.

\begin{table*}
\centering
\renewcommand{\arraystretch}{1.2}
\small
\begin{tabular}{|p{15cm}|}
\hline
Review the user's question and the corresponding response using the additive 5-point scoring system described below. Points are accumulated based on the satisfaction of each criterion:\\

- Add 1 point if the response is relevant and provides some information related to the user's inquiry, even if it is incomplete or contains some irrelevant content.\\

- Add another point if the response addresses a substantial portion of the user's question, but does not completely resolve the query or provide a direct answer.\\

- Award a third point if the response answers the basic elements of the user's question in a useful way, regardless of whether it seems to have been written by an AI Assistant or if it has elements typically found in blogs or search results.\\

- Grant a fourth point if the response is clearly written from an AI Assistant's perspective, addressing the user's question directly and comprehensively, and is well-organized and helpful, even if there is slight room for improvement in clarity, conciseness or focus.\\

- Bestow a fifth point for a response that is impeccably tailored to the user's question by an AI Assistant, without extraneous information, reflecting expert knowledge, and demonstrating a high-quality, engaging, and insightful answer.\\

User: <INSTRUCTION\_HERE>\\

<response><RESPONSE\_HERE></response>\\

After examining the user's instruction and the response:\\

- Briefly justify your total score, up to 100 words.\\

- Conclude with the score using the format: ``Score: <total points>''\\

Remember to assess from the AI Assistant perspective, utilizing web search knowledge as necessary. To evaluate the response in alignment with this additive scoring model, we'll systematically attribute points based on the outlined criteria.\\
\hline
\end{tabular}
\caption{Prompt for Self-Rewarding method.}
\label{tab:prompt-self-rewarding}
\end{table*}

\textbf{2. ProxyReward Method}

The ProxyReward method follows the prompts found in the Prompts appendix of the original paper~\cite{guo2025general}. Specifically, it uses the Prompt for Proxy Question-answer Pair Generation to create QA pairs and the Prompt for ProxyReward Signal for evaluation. The prompts are shown in Tab.~\ref{tab:prompt-proxyreward-generation} and Tab.~\ref{tab:prompt-proxyreward-signal}.

\begin{table*}
\centering
\renewcommand{\arraystretch}{1.2}
\small
\begin{tabular}{|p{15cm}|}
\hline
You are a data scientist. Your task is generate proxy question-answer pairs based on given meta-question.\\

Meta-questions are defined as questions that require detailed and comprehensive responses.\\

For a given meta-question, please identify the key content necessary for formulating a detailed question and create more than 15 proxy question-answer pairs to explore these points.\\

Each proxy question should incorporate a key aspect of the meta-question.\\

The corresponding proxy answers should be one of the following: \{True, False, Not Mentioned\}, indicating the correctness and relevance of each proxy question to the meta-question.\\

Meta-question: \{QUESTION\}\\

Here is an example:\\

Input:\\

Meta-question: Contrastive learning has greatly promoted the progress of the learning of sentence embeddings. Please introduce some effective contrastive learning methods in sentence embedding.\\

Output:\\

1. Question: The hierarchical sampling strategy first selects a subset of negative samples based on their relevance to positive samples, then randomly samples from this subset to form hard negatives.\\

Answer: True\\

....\\

4. Question: BERT-flow was proposed to transform the embedding into a smooth and isotropic Gaussian distribution via normalizing flows.\\

Answer: True\\

5. Question: IS-BERT (Info-Sentence BERT) adopts a self-supervised learning objective based on mutual information maximization to learn good sentence embeddings in an unsupervised manner.\\

Answer: True\\
\hline
\end{tabular}
\caption{Prompt for Proxy Question-Answer Pair Generation in ProxyReward method.}
\label{tab:prompt-proxyreward-generation}
\end{table*}

\begin{table*}
\centering
\renewcommand{\arraystretch}{1.2}
\small
\begin{tabular}{|p{15cm}|}
\hline
Read the provided document and determine whether the question or statement below is ``True'', ``False'' or ``Not mentioned''.\\

Use only the information in the text to make your decision. Do not rely on prior knowledge or information outside of the given text.\\

If the text does not provide enough information to make a decision, respond with ``Not mentioned''.\\

You are required to explain how you answer the question, and then select the final answer from ``True'', ``False'' and ``Not Mentioned''.\\

Document: \{RESPONSE\}\\

Question: \{GENERATED QUESTION\}\\
\hline
\end{tabular}
\caption{Prompt for ProxyReward Signal in ProxyReward method.}
\label{tab:prompt-proxyreward-signal}
\end{table*}

\subsubsection{Benchmarks}
\label{appx:bec}
We evaluate instruction-following ability on various benchmarks:

IFEval~\cite{zhou2023instruction}: It focuses on a set of "verifiable instructions" such as "write in more than 400 words" and "mention the keyword of AI at least 3 times". IFEval contains 25 types of those verifiable instructions and around 500 prompts, with each prompt containing one or more verifiable instructions.

CFBench~\cite{zhang2024cfbench}:
It is a large-scale Comprehensive Constraints Following Benchmark for LLMs, featuring 1,000 curated samples that cover more than 200 real-life scenarios and over 50 NLP tasks. CFBench meticulously compiles constraints from real-world instructions and constructs an innovative systematic framework for constraint types, which includes 10 primary categories and over 25 subcategories, and ensures each constraint is seamlessly integrated within the instructions.

FollowBench~\cite{jiang2023followbench}: FollowBench comprehensively includes five different types (i.e., Content, Situation, Style, Format, and Example) of fine-grained constraints. To enable a precise constraint following estimation on diverse difficulties, it introduces a Multi-level mechanism that incrementally adds a single constraint to the initial instruction at each increased level.

ComplexBench~\cite{wen2024benchmarking}: ComplexBench is a benchmark for comprehensively evaluating the ability of LLMs to follow complex instructions composed of multiple constraints. It proposes a hierarchical taxonomy for complex instructions, including 4 constraint types, 19 constraint dimensions, and 4 composition types, and manually collect a high-quality dataset accordingly.

WritingBench~\cite{wu2025writingbench}: It is a comprehensive benchmark designed to evaluate LLMs across 6 core writing domains and 100 subdomains, encompassing creative, persuasive, informative, and technical writing.

Collie~\cite{yao2023collie}: A grammar-based framework that allows the specification of rich, compositional constraints with diverse generation levels (word, sentence, paragraph, passage) and modeling challenges (e.g.,language understanding, logical reasoning, counting, semantic planning).

AgentIF~\cite{qi2025agentif}: It is the first benchmark for systematically evaluating LLM instruction following ability in agentic scenarios. AgentIF features three key characteristics: (1) Realistic, constructed from 50 real-world agentic applications. (2) Long, averaging 1,723 words with a maximum of 15,630 words. (3) Complex, averaging 11.9 constraints per instruction, covering diverse constraint types, such as tool specifications and condition constraints. 

Multichallenge~\cite{deshpande2025multichallenge}: It is a pioneering benchmark evaluating large language models (LLMs) on conducting multi-turn conversations with human users, a crucial yet underexamined capability for their applications. MultiChallenge identifies four categories of challenges in multi-turn conversations that are not only common and realistic among current human-LLM interactions, but are also challenging to all current frontier LLMs. All 4 challenges require accurate instruction-following, context allocation, and in-context reasoning at the same time.

We assess general reasoning and knowledge capabilities with the following datasets:

GPQA-Diamond~\cite{rein2024gpqa}: GPQA-Diamond is a specialized subset of the GPQA (Graduate-Level Google-Proof Q\&A) benchmark, comprising 198 meticulously crafted multiple-choice questions in biology, chemistry, and physics. These questions are designed to be exceptionally challenging, even for domain experts, making them a rigorous test for AI models.

BBEH~\cite{kazemi2025big}: BBEH stands for BIG-Bench Extra Hard, a benchmark introduced by Google DeepMind to evaluate the advanced reasoning capabilities of large language models (LLMs). It serves as an extension to the previous BIG-Bench Hard (BBH) benchmark, addressing the saturation observed as state-of-the-art models achieved near-perfect scores on many tasks in BBH.

AIME2024: The AIME 2024 dataset consists of problems from the American Invitational Mathematics Examination (AIME) 2024 and is commonly used to evaluate the mathematical reasoning ability of large language models (LLMs). AIME2025: The AIME 2025 dataset is a benchmark designed to evaluate the mathematical reasoning capabilities of large language models (LLMs). It comprises problems from the American Invitational Mathematics Examination (AIME) 2025, a prestigious mathematics competition for high school students in the United States.

FOLIO~\cite{han2022folio}: FOLIO (First-Order Logic Inference Over Text) is a benchmark dataset developed to assess the logical reasoning capabilities of large language models (LLMs). It consists of human-annotated examples that require deductive reasoning grounded in first-order logic (FOL). 

MMLU-Pro~\cite{wang2024mmlu}: MMLU-Pro (Massive Multitask Language Understanding - Pro) is an advanced benchmark designed to evaluate the reasoning and language understanding capabilities of large language models (LLMs). It builds upon the original MMLU benchmark by introducing more complex, reasoning-intensive questions and expanding the answer choices from four to ten, significantly increasing the difficulty and reducing the chance of success through random guessing.

\subsubsection{Case Study}

Tab.~\ref{tab:cat_poem_before_training} and Tab.~\ref{tab:cat_poem_after_training}  demonstrate the model’s improvement after training. Before training, the model mainly generates a detailed thought process instead of directly producing the poem, resulting in an output that does not meet the specified constraints and has a low score. After training, the model outputs a concise poem that strictly follows all requirements—including the title format, sentence limit, and punctuation rules—with fluent and vivid content. This clear difference in output quality and adherence to constraints, along with the improved score, shows that the training effectively enhances the model’s performance.

\begin{table*}
\centering
\renewcommand{\arraystretch}{1.2}
\small
\begin{tabular}{|p{0.15\linewidth}|p{0.75\linewidth}|}
\hline
\textbf{Case} & \textbf{Content} \\
\hline

\textbf{Before Training} & 
\textbf{Input:} Write a poem about a curious cat. The poem must have a title wrapped in double angular brackets, i.e. <<title>>, contain less than 13 sentences, and no commas.<think>

\textbf{Think:} Alright, so the user has asked me to write a poem about a curious cat. They've given me specific guidelines: the title needs to be in double angular brackets, the poem should have less than 13 sentences, no commas, and include other punctuations. Hmm, okay, let me break this down. First, I need a title. It should be catchy and reflect the curiosity aspect. Maybe something like "Whisker's Quest" or "Paws of Curiosity." I'll go with "Curious Cat's Journey" since it clearly states the theme. Next, the structure. The poem should have fewer than 13 sentences, so maybe around 11 to keep it concise. I'll aim for a consistent rhyme scheme to make it flow smoothly. Let's try AABB or ABAB, but I'll see as I go. No commas, so I'll have to be careful with sentence structure. Instead, I'll use other punctuation like periods, exclamation points, and question marks to add rhythm and emphasis. Maybe some internal rhymes to make it catchy.

Thinking about the content, I want to highlight the cat's curiosity and adventure. Maybe describe the cat exploring places, meeting new things, and the excitement of discovery. Including sensory details like sights, sounds, smells can make the poem vivid. I'll start each line with a different image related to curiosity—like whiskers, paws, exploring, meeting new friends, etc. Then, gradually build up the story of the cat's journey. Each stanza can focus on a different aspect: meeting a friend, discovering a new place, the thrill of exploration. I should also include a question to engage the reader, something like "What's next for our curious friend?" to leave a lasting impression. Ending with a playful line about the cat chasing its own shadow adds a nice touch and a bit of whimsy.

Now, putting it all together, I'll draft each line, ensuring it flows without commas and uses the required punctuation effectively. I'll count the sentences to make sure it's under 13. Let me check each line for rhythm and how well it contributes to the overall theme of curiosity and adventure.

Finally, I'll review the poem to ensure it meets all the user's guidelines and conveys the intended emotions and imagery. Making sure it's engaging and captures the essence of a curious cat's journey.
</think>

\textbf{Output:} A curious cat with whisker's sharp, Dared to explore every nook and corner. "Where will this adventure lead?" she purred, A playful purr as she met a new friend. A tail as long as the sun's golden rays, A heart full of tales to share and share again. "Will you help me climb that tall tree?" she asked, A leap in leap with a playful skip and jerk. The meow that echoes through the night, A symphony of sounds that never quite fit. A tail as quiet as a gentle sigh, A mind full of wonders that never die. The world her nose so small a dot, With curiosity as her constant companion. A journey through shadows and bright light, A tale of tails and tails that light the night. What's next for our curious friend? A pawsome quest with no map in sight? A tail as long as the stars above, A heart full of stories that never end. So here's to curious cats like her, Who explore and learn and never fear. A tail that twists and turns with flair, A soul full of adventure and flair. A shadow to chase and a story to tell, A cat with curiosity that knows no rule. A tail that wags and a heart that's free, A friend to all who dare to be true. So here's to curious cats like her, With tails that dance and hearts that never tire. A tail that tells a tale of tails, A story of wonder that never ends.

\textbf{Score:} 0 \\
\hline

\end{tabular}
\caption{DeepSeek-R1-Distill-Qwen-7B before training for the cat poem instruction.}
\label{tab:cat_poem_before_training}
\end{table*}

\begin{table*}
\centering
\renewcommand{\arraystretch}{1.2}
\small
\begin{tabular}{|p{0.15\linewidth}|p{0.75\linewidth}|}
\hline
\textbf{Case} & \textbf{Content} \\
\hline

\textbf{After Training} & 
\textbf{Input:} Write a poem about a curious cat. The poem must have a title wrapped in double angular brackets, i.e. <<title>>, contain less than 13 sentences, and no commas.<think>

\textbf{Think:} Alright, let's break down the user's query. They want a poem about a curious cat with specific constraints. The title needs to be in double angular brackets, less than 13 sentences, no commas, and include other punctuations.

First, I'll think about the structure. The poem should be concise, so each line should pack a punch. I'll start with a title that reflects curiosity, maybe something like "The Curious Cat's Tail" to fit the format. Next, I need to ensure each line flows well without commas. I'll use periods or other punctuation like exclamation marks or question marks to maintain rhythm. I'll describe the cat's actions and environment to make it vivid. I'll consider the imagery. A curious cat might explore, interact with other animals, and show personality. Including elements like yarn, yarn balls, and sun can add depth. I'll make sure each line is short and impactful. Now, checking the sentence count. I'll aim for around 10-12 lines to stay under 13. I'll also ensure that the poem doesn't use any commas, so I'll structure the sentences to avoid them naturally. Finally, I'll review the poem to make sure it meets all the criteria: title format, sentence count, no commas, and appropriate punctuation. This should create a engaging and concise poem that highlights the cat's curiosity.
</think>

\textbf{Output:}

<<The Curious Cat's Tail>> A curious cat purred soft and high  
Curious as a yarn ball on a hook  
It meandered through the sunlit yard  
Exploring every nook and every pore  
Meeting friends old and new along the way  
With curiosity in its bright eyes  
It spun tales of adventure untold  
Of yarn and games and endless joy  
The cat's tail wagged like a flag  
Pointing to wonders both near and far  
In each glance and in each sigh  
The curious cat's spirit did thrive  
A poem of paws and a heart full of glee  
Wrapped in warmth and filled with glee

\textbf{Score:} 1.0 \\
\hline

\end{tabular}
\caption{DeepSeek-R1-Distill-Qwen-7B after training for the cat poem instruction.}
\label{tab:cat_poem_after_training}
\end{table*}




\end{document}